\documentclass{article}

\usepackage{graphicx}
\usepackage{subcaption}
\usepackage{romannum}
\usepackage{rotating}
\usepackage{graphicx}%
\usepackage{multirow}%
\usepackage{amsmath,amssymb,amsfonts}%
\usepackage{amsthm}%
\usepackage[title]{appendix}%
\usepackage{xcolor}%
\usepackage{textcomp}%
\usepackage{manyfoot}%
\usepackage{makecell}
\usepackage[ruled,vlined]{algorithm2e}
\usepackage{algpseudocode}%
\usepackage{listings}%
\usepackage{pgfplots}
\pgfplotsset{compat=1.18} 
\usepackage{array}
\usepackage{booktabs}
\usepackage{float}
\usepackage{hyperref}

\theoremstyle{thmstyleone}%

\theoremstyle{thmstyletwo}%

\theoremstyle{thmstylethree}%
\usepackage{authblk}
\usepackage{caption}

\title{A Two-Stage Forecasting System for CPU Workload Prediction in Private Clouds}

\author[1]{Ashir Javeed}
\author[1]{Anton Borg\footnote{Corresponding author. Email: \texttt{anton.borg@bth.se}}}
\author[1]{H{\aa}kan Grahn}
\author[1]{Lars Lundberg}
\author[2]{Dhyey Patel}
\author[2]{Sogand Shirinbab}

\affil[1]{Department of Computer Science, Blekinge Institute of Technology, 371 79, Sweden}
\affil[2]{Ericsson AB, 371 33 Karlskrona, Sweden}

\date{}
\begin{document}

\maketitle

\begin{abstract}
Accurate cloud resource forecasting is essential for proactive resource provisioning, maintaining Quality of Service (QoS), and reducing operational costs in dynamic cloud environments. The existing forecasting approaches predominantly estimate future CPU workload directly from historical resource traces, which often overlook the relationship between customer service demand and subsequent resource consumption. This study proposes a two-stage integrated forecasting model that explicitly models this dependency by first forecasting customer service requests, expressed as Transactions Per Second (TPS), and subsequently estimating future CPU workload from the TPS forecast. Both the forecasting component and resource prediction component employed the XGBoost model within a cascaded learning architecture, complemented by adaptive online retraining using an expanding-window strategy to address concept drift in continuously evolving cloud workloads. The proposed work was evaluated using real-world traces collected from a private cloud environment comprising ten applications. Experimental results demonstrate robust forecasting performance by achieving Symmetric Mean Absolute Percentage Error (SMAPE) below $7\%$ for most applications, with the best-performing application achieving an MAE of $0.7372$, RMSE of $1.1866$, SMAPE of $3.57\%$, and an $R^2$ of $0.9185$. Horizon-wise drift analysis confirmed stable recursive forecasting behavior with controlled error accumulation across a 60-step prediction horizon. Compared with the conventional direct CPU forecasting method, the newly developed two-stage integrated model gives improved forecasting robustness, computational efficiency, and interpretability, making it well-suited for proactive resource management and intelligent auto-scaling in cloud computing environments.
\end{abstract}


\section{Introduction}\label{sec1}
Cloud computing has experienced tremendous growth in recent years, driven by lower operating costs and increased productivity~\cite{muller2015benefits}. The recent advancements in artificial intelligence (AI) and Big data also contribute to the steadily rising demand for cloud computing, as these technologies require significant computational resources~\cite{marr2020tech}. To maintain the Quality of Service (QoS), cloud service providers should allocate and manage computing resources efficiently while respecting energy and sustainability factors. Over-provisioning of cloud resources costs in terms of energy waste, and under-provisioning affects QoS~\cite{al2022workload}. Therefore, cloud computing resources such as processing (CPU) power, memory, and storage should be allocated efficiently to meet cloud customers' needs and energy. Consequently, maintaining smooth cloud operations and efficiency requires accurate forecasting of future demand for cloud resources.

Cloud service providers rely on workload forecasting to proactively allocate computing resources and maintain QoS under dynamic workload conditions~\cite{patel2023integrated}. In this regard, historical data can be utilized for building a predictive model for forecasting an estimated demand of computing resources such as CPU, storage, memory, and network bandwidth for future needs. However, accurately estimating future demand for computing resources is not an easy task because workload in data centers is highly dynamic and rapidly changing for demanded resources. For instance, 5\% to 80\% variation in cloud workload is observed in Alibaba cloud datacentres~\cite{chen2018does}. Similar variation in cloud workload is also studied in this study, where data is obtained from a private cloud services provider company. Developing an efficient method that can accurately address high variance in cloud resource demand requires an in-depth understanding and knowledge of workload patterns~\cite{xu2024practice}. Through acquired knowledge, we can construct an algorithm that is robust and precise for forecasting the future workload of the cloud.       

Generally, cloud data is sequential in nature and collected over time, which turns resource forecasting into a time-series problem~\cite{villegas2018support}. Therefore, researchers have utilized statistical methods such as the ARIMA model and machine learning models, including linear regression, decision, and most recent methods based on deep learning, such as recurrent neural network (RNN) and Long Short-Term Memory (LSTM), to deal with the problem as a time series task~\cite{esposito2025forecasting, javeed2025improving}. These models successfully predict future cloud workload, but they do not account for the uncertainty of such forecasts due to limited data on cloud customer behavior. Furthermore, few studies in the literature considered the factor of services requested by the cloud customers for future workload forecasting~\cite{javeed2025cloudtech,sugan2024predictopticloud, wang2021predicting}. Additionally, forecasting resource requirements across cloud applications based on customer demand remains a challenge for cloud researchers. Moreover, the presence of high fluctuations across different services based on customers' demand at various time-steps, and their impact on cloud resources such as CPU utilization in the future, poses a significant challenge for accurate forecasting of future resource (CPU). Hence, there is a need for a mechanism that accounts for customer service requests at a given time step for a particular cloud application and accurately forecasts the future resources needed to handle customer demand. Existing forecasting methods directly model future CPU utilization from historical CPU traces. However, CPU consumption is primarily a downstream consequence of incoming service demand. Ignoring this causal dependency limits forecasting robustness under highly dynamic cloud workloads.

To overcome these limitations, this study proposes a two-stage integrated forecasting system for proactive CPU resource prediction in cloud environments. We build on our previous work \cite{ javeed2025cloudtech}, where customer requests were used as input features to train a machine learning model for CPU workload prediction in cloud applications. In the present study, a dedicated forecasting model is developed to predict future customer requests, which are subsequently utilized as input features for downstream CPU workload estimation. The integration of the CPU model with the forecasting model provides a comprehensive understanding of feature (customer requests) interactions over time, thereby enhancing the model’s forecasting capabilities. Contrary to computationally expensive deep sequential forecasting architectures such as LSTM and Transformer-based forecasting models~\cite{esposito2025forecasting}, the proposed XGBoost-based framework supports lightweight adaptive learning with lower retraining overhead and faster inference for streaming cloud environments. The contributions and findings of our work are summarised as follows:

\begin{itemize}
\item The study presents an integrated forecasting system that models customer service requests and downstream CPU utilization in two sequential learning stages.

\item Developed a TPS-driven forecasting mechanism that captures workload dynamics prior to CPU estimation for proactive cloud resource management.

\item Adaptive online retraining and rolling maximum (Rolling\_Max) smoothing were integrated into the proposed system to enhance forecasting stability under highly dynamic cloud computing workloads.

\item Demonstrated computationally efficient forecasting using XGBoost-based learning suitable for lightweight streaming deployment.

\item Validated the proposed system using real-world private cloud traces and showed improved forecasting accuracy and robustness compared with conventional forecasting approaches.

\end{itemize}

The rest of the study is structured as follows. Section~\ref{sec:relatedwork} presents related work. Section~\ref{sec:Methodology} presents the methodology. The experimental results and the statistical analysis are elaborated in Section~\ref{sec:experiment}. Section~\ref{sec:discussion} provides a detailed discussion of the given work. Finally, Section~\ref{sec:conclusion} presents the conclusion.

\section{Related Work}
\label{sec:relatedwork}
Resource forecasting in cloud data center is a critical component for efficient resource management; therefore, significant research attention has been devoted in the literature to address this problem. Researchers have employed different methodologies based on statistics, conventional machine learning and state-of-the-art novel techniques for accurate resource forecasting. Statistical models are primarily based on mathematical principles that help to identify the trends and patterns in the data. However, these models struggle to capture the patterns when the data contains complex and nonlinear relationships. Consequently, accurate resource forecasting in cloud environments becomes challenging when relying on traditional statistical models such as ARIMA~\cite{calheiros2014workload}. ARIMA is based on AutoRegressive (AR) and Moving Average (MA) principles, which are effective in capturing stationary trends and seasonality. Therefore, ARIMA has limited capability in capturing nonlinear relationships and complex patterns in cloud resource forecasting data~\cite{kontopoulou2023review}.

On the other hand, ML models demonstrate superior capability in learning nonlinear workload patterns for efficient cloud resource forecasting over statistical models. ML models are sometimes computationally expensive because they require feature engineering~\cite{zheng2018feature} to learn complex pattern in the data and hyperparameter tuning to improve performance~\cite{albahli2024efficient}.  For instance, in literature different ML models are utilized for resource prediction and forecasting in cloud environment such as random forest~\cite{valarmathi2021resource}, XGBoost~\cite{rathee2026optimizing}, K-nearest neighbors regression~\cite{martinez2018dealing}, support vector regression~\cite{sripathi2026proactive}, and ensemble~\cite{banerjee2021efficient, javeed2025improving}. Although machine learning approaches improve forecasting capability under nonlinear cloud workload conditions, their performance often depends heavily on feature engineering and hyperparameter optimization. Furthermore, many ML-based forecasting systems primarily focus on prediction accuracy, offering limited adaptability to evolving workload dynamics in streaming cloud environments.

A. Rathee et al. proposed an integrated model (XGBoost and Red Fox Optimization Algorithm (RFOA)) for optimized resource allocation in a cloud environment. In their proposed model, RFOA algorithm was deployed to optimize the hyperparameters of XGboost model. The performance of the proposed model was compared with the baseline and genetic algorithm (GA)-based models. Experimental results displayed that proposed model achieved superior performance in term of accuracy and resource utilization while reduced computational cost and improve reliability of cloud services~\cite{rathee2026optimizing}. G. Sripathi and D. A. Khan presented a proactive resource allocation framework using support vector machine (SVM) regression model for cloud computing environment. The developed framework employed workload and geographic information to predict QoS requirements and dynamic allocated resources across geo-distributed cloud locations. The proposed method helps in efficient resource utilization, reduced overprovisioning and underutilization, and enhanced overall system responsiveness and QoS performance. Experimental results demonstrated significant improvements in prediction accuracy, precision, recall, and F1-score compared to existing methods~\cite{sripathi2026proactive}.

Furthermore, N. Roy et al. presented an autoscaling predictive method for workload forecasting in cloud environment for improved Quality of Service (QoS). Their predictive model is based on a second-order autoregressive moving average method that accurately forecasts future workload and satisfies the application QoS while keeping operational costs low~\cite{roy2011efficient}. Y. C. Chang et al. developed an algorithm based on Recurrent Neural Network (RNN) model for efficient workload forecasting in cloud computing servers. Their proposed method helps to proactive resource allocation through predicting future workload in advance that improve resource management in the cloud~\cite{chang2013predictive}. Z. Chen et al. proposed a hybrid Fuzzy Neural Network (FNN) with ensemble modelling approach for accurate forecasting of future workload in cloud. The developed model combined fuzzy logic and ensemble learning for better capturing the complex workload patterns in the cloud environment that helps in accurate forecasting of resources and efficient cloud resource management~\cite{chen2015self}. Z. Chen et al. developed an algorithm, namely the Prediction Algorithm for Cloud Computing Workloads, based on a deep learning technique for cloud workload forecasting. The performance of the conventional RNN model was improved by integrating a Top-Sparse Autoencoder (TSA) with a Gated Recurrent Unit (GRU). TSA helps by extracting the useful workload patterns from the data, and the future workload is accurately predicted with GRU. Their proposed algorithm was evaluated on the real-world workload traces from Google and Alibaba Cloud. The experimental results showed that the proposed algorithm outperformed the traditional RNN model in workload forecasting accuracy~\cite{chen2019towards}. Although deep learning models improve nonlinear sequence modeling capability, they typically require computationally expensive retraining, large-scale datasets, and substantial hardware resources, limiting their practicality for lightweight adaptive cloud deployment.

In 2021, M. A. Razz et al. designed a hybrid auto-scaling framework for workload forecasting in cloud computing environment. The developed framework is based on an ensemble technique that employs several ML models, such as a support vector machine, a decision tree, a random forest, and a K-nearest neighbors. To identify the dynamic workload variations and better Quality of Service (QoS), their framework introduced a burst-aware auto-scaling mechanism that improved resource utilization via vertical and horizontal scaling and lowered computational overhead. The proposed framework also analyzed the service requests by the cloud users to tackle the sudden workload burst efficiently~\cite{razzaq2021hybrid}. S. Karimunnisa et al. proposed a workload forecasting model by employing deep learning techniques to tackle dynamic workload patterns in cloud computing. Their proposed model integrated Deep Belief Network (DBN) with the Lion Algorithm (LA) that optimized the hidden neuron of the DBN model for improve prediction accuracy. The study experimental results demonstrated that proposed model outperformed existing state-of-the-art forecasting approaches~\cite{karimunnisa2025novel}.

Recently in 2025, A. Rossi et al. presented a Bayesian deep learning (BDL) model that forecast future workload of a cloud to improve better resource allocation and QoS reliability in cloud computing environments. The presented model introduced univariate and bivariate prediction models using Hybrid Bayesian Neural Networks (BNNs) and probabilistic Long Short-Term Memory (LSTM) networks to forecast multiple cloud resources. The developed method also employed transfer learning techniques to improve adaptability across different cloud data centres. Experimental evaluation done using Google and Alibaba cloud datasets demonstrated improved forecasting accuracy and better service-level performance compared to traditional forecasting models~\cite{rossi2025forecasting}. Table~\ref{tab:literature_summary} presents the summary of studies discussed in the literature that consists of study references, key contribution, methodology and findings of the given study.

\begin{sidewaystable*}[p]
\centering
\footnotesize
\caption{Comparative summary of cloud resource forecasting studies}
\label{tab:literature_summary}

\renewcommand{\arraystretch}{1.15}

\begin{tabular}{
>{\raggedright\arraybackslash}p{1.1cm}
>{\raggedright\arraybackslash}p{2.9cm}
>{\raggedright\arraybackslash}p{3.4cm}
>{\raggedright\arraybackslash}p{1.4cm}
>{\raggedright\arraybackslash}p{3.0cm}
>{\raggedright\arraybackslash}p{6.0cm}
}
\toprule

Ref. & Methodology & Forecast Target & Adaptive Learning & Dataset / Environment & Key Limitation \\

\midrule

~\cite{javeed2025improving}
& Ensemble Learning
& CPU workload prediction
& Yes
& Cloud applications
& Direct CPU forecasting without explicit workload causality modeling. \\

~\cite{calheiros2014workload}
& Statistical (ARIMA)
& Cloud workload
& No
& Cloud workload traces
& Effective for stationary trends but weak for nonlinear and dynamic cloud workloads. \\

~\cite{valarmathi2021resource}
& Random Forest
& Resource utilization
& No
& Cloud resource traces
& Requires feature engineering and lacks online adaptability. \\

~\cite{rathee2026optimizing}
& XGBoost + RFOA
& Resource allocation
& Partial
& Cloud computing environment
& Hyperparameter optimization adds computational overhead. \\

~\cite{martinez2018dealing}
& KNN Regression
& Workload prediction
& No
& Dynamic cloud workloads
& Sensitive to neighborhood selection and workload variance. \\

~\cite{sripathi2026proactive}
& SVM Regression
& QoS-aware resource allocation
& Partial
& Geo-distributed cloud environment
& Computational overhead increases with workload scale and feature complexity. \\

~\cite{banerjee2021efficient}
& Ensemble ML
& Cloud resource forecasting
& No
& Cloud workload traces
& Improved accuracy but increased model complexity and retraining cost. \\

~\cite{roy2011efficient}
& ARIMA-based forecasting
& Autoscaling workload
& No
& Cloud application traces
& Limited support for rapidly fluctuating workloads. \\

~\cite{chang2013predictive}
& Deep Learning (RNN)
& Workload forecasting
& No
& Cloud server workloads
& High retraining cost and limited long-term dependency learning. \\

~\cite{chen2015self}
& Hybrid FNN + Ensemble
& Cloud workload
& No
& Cloud computing environment
& Increased architectural complexity and computational cost. \\

~\cite{chen2019towards}
& TSA + GRU
& Cloud workload prediction
& No
& Google and Alibaba traces
& Requires large-scale training data and expensive retraining. \\

~\cite{razzaq2021hybrid}
& Hybrid Ensemble + Burst-aware Scaling
& Dynamic workload forecasting
& Partial
& Cloud autoscaling environment
& Ensemble framework increases deployment complexity in streaming environments. \\

~\cite{karimunnisa2025novel}
& DBN + Lion Algorithm
& Workload prediction
& No
& Cloud workload traces
& Deep learning optimization introduces high computational overhead. \\

~\cite{rossi2025forecasting}
& Bayesian DL (BNN + Probabilistic LSTM)
& Multi-resource forecasting
& Partial
& Google and Alibaba datasets
& High computational complexity and expensive adaptive retraining. \\

\midrule

Current Study
& two-stage integrated XGBoost forecasting
& TPS + CPU forecasting
& Yes
& Private cloud environment
& Adaptive forecasting with explicit workload-to-CPU causal modeling and reduced retraining overhead. \\

\bottomrule
\end{tabular}
\end{sidewaystable*}

Most current cloud forecasting research uses historical resource traces to directly estimate future CPU consumption. However, inbound workload intensity and customer service demand are intrinsically what cause CPU consumption. As a result, in cloud settings, direct forecasting methods might not adequately capture the causal relationship between workload evolution and downstream infrastructure response.

Although significant progress has been made in cloud resource forecasting, several challenges remain: statistical models often struggle to capture complex, nonlinear workload patterns, and machine learning and deep learning approaches typically require substantial computational resources, extensive hyperparameter tuning, and large-scale training data. In addition, many existing methods primarily focus on improving prediction accuracy while inadequately addressing real-time adaptability and efficient resource management in cloud environments. Therefore, there is a need for a simple, efficient forecasting system that accurately predicts cloud resource (CPU)  demand in a private cloud system while maintaining computational efficiency.
In contrast to traditional direct forecasting techniques, which estimate future CPU utilization solely from historical CPU traces, the proposed two-stage integrated system decomposes the forecasting problem into two sequential learning stages: customer request forecasting and downstream CPU estimation. The proposed cascaded formulation explicitly models the causal relationship between incoming service demand and application-level resource consumption (CPU). Because CPU utilization is mostly determined by workload intensity, forecasting TPS before CPU prediction allows the system to detect workload evolution sooner than direct resource-forecasting approaches. Furthermore, decoupling workload dynamics from infrastructure response simplifies forecasting, increases interpretability, and enhances flexibility in highly dynamic private cloud environments.

\section{Methodology}
\label{sec:Methodology}

This section presents the proposed two-stage integrated forecasting system for proactive CPU resource prediction in private cloud environments. The system integrates multivariate Transactions Per Second (TPS) forecasting with a downstream CPU estimation model using a unified machine learning pipeline. Unlike traditional approaches that directly map historical CPU usage to future demand, the proposed method explicitly models customer service requests  as an intermediate representation, enabling earlier and more robust prediction of CPU utilization under dynamic workload conditions.

\begin{figure*}[hbt]
	\centering
	\includegraphics[width=0.75\linewidth]{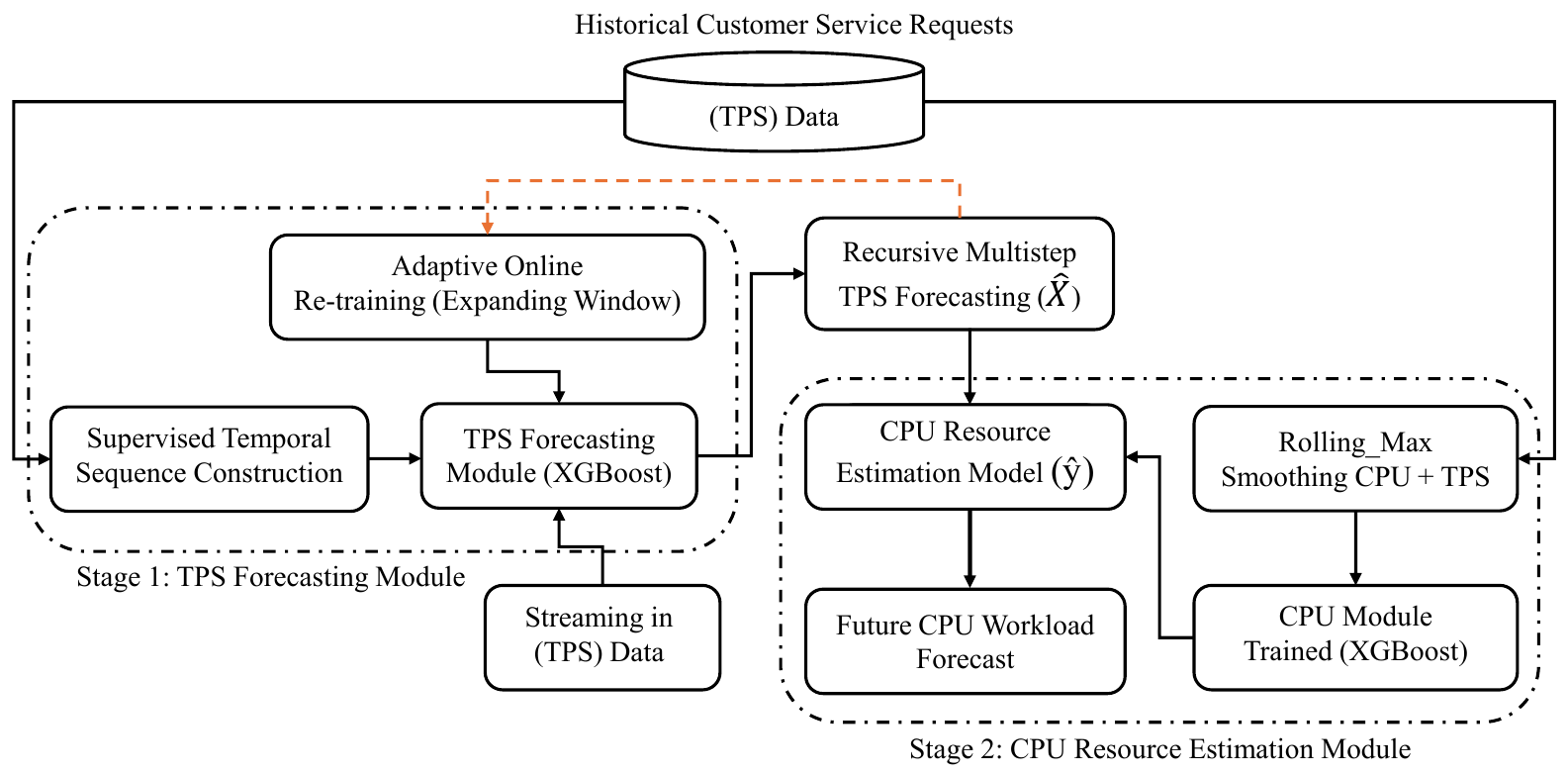}
	\caption{Architecture of the proposed two-stage integrated forecasting system with adaptive expanding-window retraining for proactive CPU resource prediction in private cloud.}
	\label{fig:Proposed_System}
\end{figure*}

The architecture consists of two stages: (i) TPS Forecasting Module and (ii) CPU Resource Estimation Module, as illustrated in Figure~\ref{fig:Proposed_System}. The system operates in a walk-forward streaming setting with periodic retraining using an expanding-window strategy to handle concept drift in cloud environments.

\subsection{Problem Formulation}

Let the multivariate TPS observation at time step $o$ be defined as:
\begin{equation}
X_o = [x_o^{(1)}, x_o^{(2)}, \ldots, x_o^{(m)}],
\label{lab:tps_features}
\end{equation}
where $m$ denotes the number of TPS-related features.

The corresponding CPU utilization is represented as:
\begin{equation}
y_o \in \mathbb{R}.
\end{equation}

The objective is to predict future CPU utilization over horizon $h$ using historical TPS observations.

\subsection{Stage 1: TPS Forecasting Module}

A supervised learning dataset is constructed using a sliding historical window of size $hw = 512$:

\begin{equation}
\mathbf{S}_o =
[\mathbf{x}_{o-hw+1}, \ldots, \mathbf{x}_o] \in \mathbb{R}^{hw \times m}.
\end{equation}

A multi-output XGBoost regression model learns the mapping:
\begin{equation}
f_{\theta}(\mathbf{S}_o) = X_{o+1}.
\end{equation}

Multi-step forecasting is performed recursively:
\begin{equation}
\hat{X}_{o+i} = f_{\theta}(\hat{\mathbf{S}}_{o+i-1}), \quad i = 1,2,\ldots,h,
\end{equation}

The resulting TPS forecast trajectory is:
\begin{equation}
\hat{X}_{o:o+h} =
[\hat{X}_{o+1}, \ldots, \hat{X}_{o+h}].
\end{equation}

\subsection{Stage 2: CPU Resource Estimation Module}

The CPU estimation model is trained independently on original TPS feature vectors, as given in Equation~\ref{lab:tps_features}. When training the CPU target value is defined using a rolling maximum window:
\begin{equation}
\tilde{y}_o =
\max (y_{o+1}, y_{o+2}, \ldots, y_{o+\beta}),
\end{equation}

which captures near-future peak CPU demand while reducing sensitivity to short-term fluctuations.

The CPU estimation model is trained as:
\begin{equation}
g_{\phi}(X_o) = \tilde{y}_o.
\end{equation}

The training dataset is defined as:
\begin{equation}
\mathcal{D}_{CPU} = \{(X_o, \tilde{y}_o)\}_{o=1}^{N}.
\end{equation}

\subsection{Integrated Prediction Mechanism}

During inference, the CPU model directly consumes the forecasted TPS feature  from the first stage, without using recursive state representations:

\begin{equation}
\hat{y}_{o+i} = g_{\phi}(\hat{X}_{o+i}), \quad i = 1,2,\ldots,h.
\end{equation}

The final CPU forecast trajectory is:
\begin{equation}
\hat{Y}_{o:o+h} =
[\hat{y}_{o+1}, \ldots, \hat{y}_{o+h}].
\end{equation}

The TPS forecasting model is periodically retrained using an expanding-window strategy after every r=1000 observations. The retraining interval (r=1000) was empirically selected to balance adaptation speed, prediction accuracy, and computational overhead. More frequent retraining increases computational cost with only marginal gains in accuracy, whereas less frequent retraining delays adaptation to changes in workload patterns. This strategy enables the model to efficiently adapt to evolving workload distributions while mitigating concept drift in streaming cloud environments.

Algorithm~\ref{alg:framework} summarizes the complete workflow of the proposed system. The algorithm first trains a multivariate TPS forecasting model using a sliding-window representation of historical workload data. It then independently trains a CPU estimation model using instantaneous TPS feature vectors paired with a forward-looking rolling maximum CPU target. During deployment, the system operates in a walk-forward manner where TPS values are recursively forecasted over the prediction horizon. These predicted TPS values are then directly passed to the CPU model to estimate future CPU utilization. Periodic retraining of the TPS model using an expanding window ensures adaptability to non-stationary workload dynamics while maintaining computational efficiency.

\begin{algorithm}[ht]
\caption{Adaptive Two-Stage TPS-Driven CPU Forecasting}
\label{alg:framework}

\KwIn{TPS data $\mathbf{X}$, CPU data $\mathbf{Y}$, window size $hw$, horizon $h$, rolling window $\beta$, retraining interval $r$}
\KwOut{Predicted CPU trajectory $\hat{\mathbf{Y}}_{o:o+h}$}

Construct TPS training set $\mathcal{D}_{TPS}=\{(\mathbf{S}_o,X_{o+1})\}$ and train TPS model $f_{\theta}$.

Construct CPU training set $\mathcal{D}_{CPU}=\{(X_o,\tilde{y}_o)\}$, where
$\tilde{y}_o=\max(y_{o+1},\ldots,y_{o+\beta})$, and train CPU model $g_{\phi}$.

\For{each time step $o$}{

Append newly observed TPS and CPU data.

\If{$o \bmod r = 0$}{
Retrain $f_{\theta}$ using the expanding-window dataset.
}

Initialize $\hat{\mathbf{S}}_o \leftarrow \mathbf{S}_o$.

\For{$i=1$ \KwTo $h$}{

$\hat{X}_{o+i}=f_{\theta}(\hat{\mathbf{S}}_{o+i-1})$

Update $\hat{\mathbf{S}}_{o+i}$ with $\hat{X}_{o+i}$

$\widehat{\tilde{y}}_{o+i}=g_{\phi}(\hat{X}_{o+i})$

}

$\hat{\mathbf{Y}}_{o:o+h}=[\widehat{\tilde{y}}_{o+1},\ldots,\widehat{\tilde{y}}_{o+h}]$

}

\Return{$\hat{\mathbf{Y}}_{o:o+h}$}

\end{algorithm}

\subsection{Dataset Collection and Environment}
The dataset used in this study was collected from a private cloud environment. Resource (CPU) utilization, along with the cloud services requested by customers, was recorded at 1-minute intervals. Data were collected for one day, resulting in a total of 1,440 samples, which was insufficient for the scope of this study. Therefore, the dataset was expanded by replicating and mirroring the original data points, increasing its size by approximately three times to obtain 8,514 samples, representing roughly six days of data.

The applications are organized into pods in the cloud computing environment, where a pod represents a group of one or more applications deployed together and managed as a single operational unit. In this study, we focused on applications whose CPU utilization is primarily influenced by user service requests. Four pods, A, B, C, and D, were selected from the private Cloud. Pods A and C each contain one application, pod B contains three ($B1$, $B2$, $B3$), and pod D contains five ($D1$–$D5$). Confidentiality agreements with the data provider prevent the disclosure of the specific definitions of certain attributes (e.g., specific service-level metrics or measurement units).  However, selected features align with those explored in similar feature-based studies (e.g., Javeed et al.~\cite{javeed2025improving}, Wang et al.~\cite{wang2018cloud}), which typically capture user service request patterns. Table~\ref{Table:DatasetDescription} summarizes the details of each pod, including the number of input features per application, and the CPU utilization statistics: range, mean, and standard deviation (SD). Moreover, the CPU utilization range in Table~\ref{Table:DatasetDescription} is multiplied by a constant factor to hide sensitive data information.

\begin{table*}[ht]
	\centering
	\caption{Description of the selected applications.}
	\label{Table:DatasetDescription}
	\begin{tabular}{@{}llllll@{}}
		\toprule
		Pod & Application & Features (TPS) & CPU Range&Mean& SD\\
		\midrule
		A
		& $A$ & 5 & 03.8342 -- 49.6500 & 30.8940 & 06.6864 \\
		\multirow{3}{*}{B}
        & $B1$ & 5 & 67.6155 -- 275.625 & 187.890 & 47.4330 \\
		& $B2$ & 5 & 04.8462 -- 12.2025 & 06.4925 & 01.5356 \\
		& $B3$ & 5 & 06.6900 -- 18.1230 & 11.1893 & 02.2032 \\

		C
		& $C$ & 5 & 02.2169 -- 61.1310 & 24.6960 & 11.4692 \\

		\multirow{5}{*}{D}

        & $D1$ & 5 & 0.04910 -- 0.08250 & 0.06660 & 0.00840 \\
		& $D2$ & 5 & 0.01940 -- 0.15590 & 0.06660 & 0.05390 \\
		& $D3$ & 5 & 00.8825 -- 02.5704 & 01.5798 & 0.68970 \\
		& $D4$ & 5 & 0.00045 -- 0.01425 & 00.0048 & 0.00530 \\
		& $D5$ & 5 & 0.00255 -- 0.01125 & 00.0053 & 0.00230 \\

		\bottomrule
	\end{tabular}
\end{table*}

\begin{figure}[ht]
	\centering
	\includegraphics[width=\linewidth]{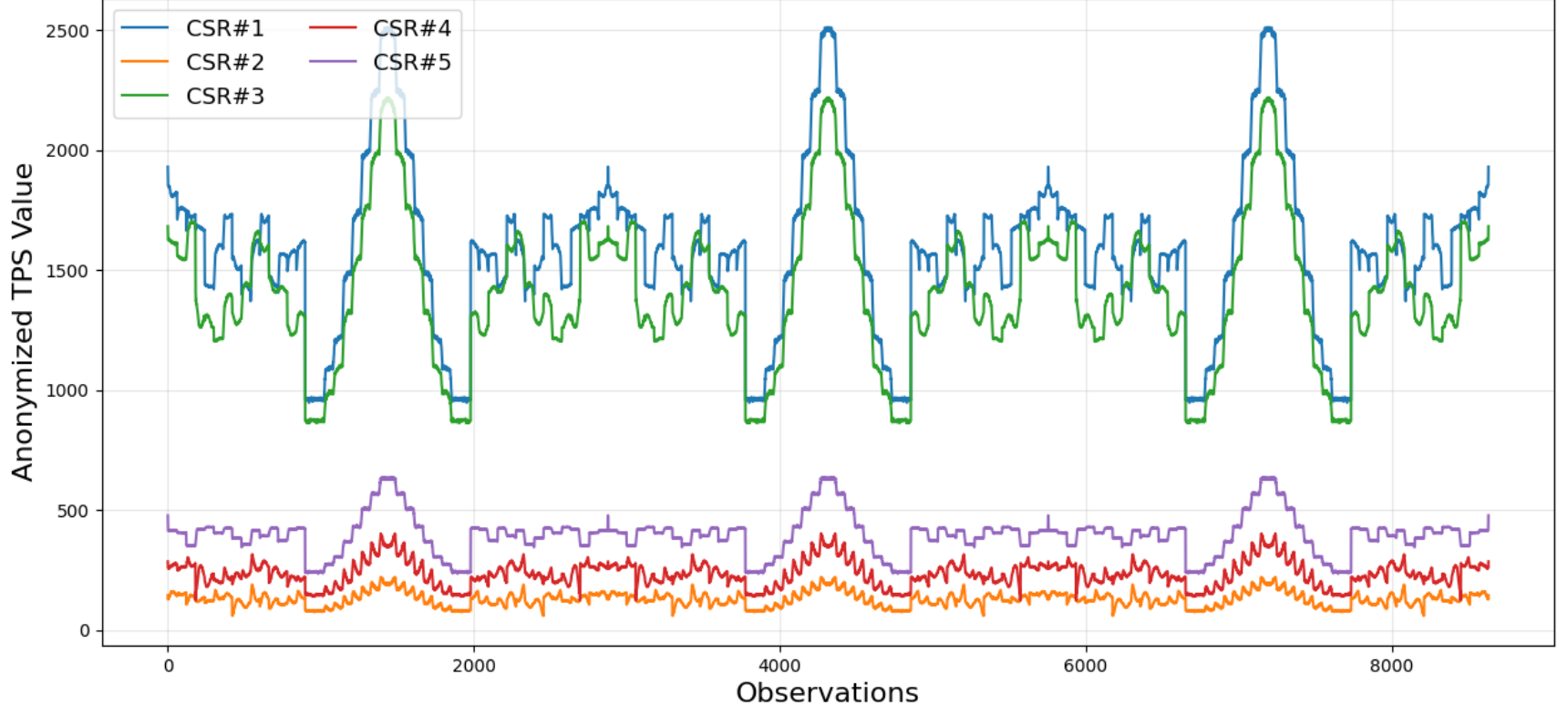}
	\caption{Flow of incoming customer service requests (TPS) to the private cloud}
	\label{fig:Incoming_Requests}
\end{figure}

The incoming service requests from users to the private cloud environment are shown in Figure~\ref{fig:Incoming_Requests}. Although timestamps are not explicitly modeled, the data is treated as a sequential time series using observation order. Therefore, the x-axis of Figure~\ref{fig:Incoming_Requests} represents the observation index corresponding to individual data samples for the given service request. The y-axis shows the number of customer service requests (TPS). The true values on the y-axis for requested services were anonymized to protect the privacy of data.

All experiments were conducted using Python (version 3.12.3) on a local machine. The details of the computing environment are summarized in Table~\ref{tab:Configuration}.

\begin{table}[ht]
    \centering
    \caption{Setup of computing environment.}
    \label{tab:Configuration}
        \begin{tabular}{@{}ll@{}}
            \toprule
            System properties & Specifications  \\
            \midrule
            CPU & Intel(R) Core i5-1145G7 @ 2.60GHz \\
            RAM & 32GB \\
            OS & Windows 10 Enterprise \\
            Storage & 475GB \\
            IDE & Visual Studio 1.92 \\
            \bottomrule
        \end{tabular}%

\end{table}

\subsection{Validation and Evaluation Metrics}
For the problem addressed in this study, we employed a walk-forward validation scheme, which is well-suited to time-dependent data and forecasting tasks that require temporal ordering to be preserved~\cite{hyndman2021forecasting}. In the walk-forward validation method, the model is iteratively trained on a historical window while evaluated on the subsequent unseen observation period. This validation method closely simulates real-world deployment conditions~\cite{bergmeir2018note}. Hence, this procedure eliminates the risk of information leakage from future data and provides a realistic estimate of predictive performance under non-stationary dynamics~\cite{lau2025fast}. Initially, the first 25\% of the dataset was used to train the two-stage forecasting system. During walk-forward validation, predictions were generated sequentially for incoming observations, while the training window was expanded and the model was retrained after every r=1000 observations. This strategy preserves temporal ordering while reducing computational overhead. To assess the feasibility of real-time deployment, the execution time of each forecasting iteration was also recorded during inference. The average runtime is computed as:

\begin{equation}
Runtime_{avg} =
\frac{1}{N}
\sum_{i=1}^{N} t_i
\label{eq:runtime}
\end{equation}

where $t_i$ represents the execution time for the $i^{th}$ forecasting iteration.

The accuracy of regression-based problems is measured in terms of error~\cite{saxena2023performance}. Therefore, we have selected several error metrics to rigorously evaluate the efficiency of the constructed integrated forecasting model, such as root mean square error (RMSE)~\cite{hoffmann2007best}, mean underestimate error (MUE)~\cite{nawrocki2024optimization}, mean overestimate error (MOE)~\cite{liao2024retrospecting}, mean absolute error (MAE), and symmetric mean absolute percentage error (SMAPE). SMAPE was included because it provides a scale-independent and symmetric evaluation of forecasting accuracy, making it suitable for dynamic cloud workloads with fluctuating CPU utilization patterns. In addition, MUE and MOE were selected to analyze asymmetric underestimation and overestimation behaviors in cloud resource provisioning scenarios. Because the CPU utilization ranges differ across the evaluated applications (e.g., $A$, $B1$, $C$, and $D1$), relative metrics such as rMUE and rMOE were employed to enable fair comparisons across workloads with different scales. In addition to error metrics, we have also employed R-squared ($R^2$) to assess how well the constructed forecast system predictions match the observed CPU load. By comparing predicted values with actual CPU load, the performance of the proposed system is evaluated while accounting for the total data variability of the cloud environment. The mathematical formulas for the selected metrics are as follows:

\begin{equation}
     RMSE = \sqrt{\frac{1}{n} \sum_{i=1}^{n} (y_i - \hat{y}_i)^2}
     \label{eq:RMSE}
\end{equation}

\begin{equation}
\mathrm{MAE} =
\frac{1}{n}
\sum_{i=1}^{n}
\left| y_i - \hat{y}_i \right|
\label{eq:MAE}
\end{equation}

\begin{equation}
SMAPE =
\frac{100}{n}
\sum_{i=1}^{n}
\frac{|y_i - \hat{y}_i|}
{\left(|y_i| + |\hat{y}_i|\right)/2}
\label{eq:smape}
\end{equation}

\begin{equation}
        rMUE = \frac{1}{n} \sum_{i=1}^{n} \frac{\max(0, y_i - \hat{y}_i)}{y_i}
	\label{eq:mue}
\end{equation}

\begin{equation}
	rMOE =  \frac{1}{n} \sum_{i=1}^{n} \frac{\max(0, \hat{y}_i - y_i)}{y_i}
	\label{eq:moe}
\end{equation}

Where $y_i$ is the actual CPU value, $\hat{y}_i$ is the predicted CPU value by the ML model, and $n$ is the number of data points. In Equation~\eqref{eq:mue}, $\max(0, y_i - \hat{y}_i)$ ensures that only the underestimation errors (where the actual value is greater than the predicted value) are considered, similarly, in Equation~\eqref{eq:moe}, $\max(0, \hat{y}_i - y_i)$ ensures that only the overestimation errors (where the predicted value is greater than the actual value) are considered.

\begin{equation}
	R^2 = 1 - \frac{\sum_{i=1}^{n} (y_i - \hat{y}_i)^2}{\sum_{i=1}^{n} (y_i - \bar{y})^2}
	\label{eq:r2}
\end{equation}

$R^{2}$ is the coefficient of determination that helps in evaluating the goodness-of-fit of the proposed forecasting system. When $R^{2}$ equals 1, it means the model perfectly predicts the CPU load without any error, explaining all the variance in the actual CPU load, and when $R^{2}$ equals 0, it means the model does not explain any of the variance in the CPU load and performs no better than simply predicting the mean of the observed CPU load. Even in some cases, $R^{2}$ can be negative ($R^{2} <0$); that means the model performs worse than predicting the mean value for all data points.

\section{Experimental Results and Analysis}
\label{sec:experiment}
This section presents a detailed description of the experiments and results conducted to evaluate the performance of the newly developed two-stage integrated system for forecasting cloud resource (CPU). Customer service requests (TPS) are used as input features to forecast future CPU workloads for given cloud applications. The experimental results are presented in several subsections. Subsection~\ref{subsection: Forecasting Model Performance} presents the overall performance of the proposed integrated forecasting model across all applications using multiple evaluation metrics. Subsection~\ref{subsection: Detailed Drift Analysis} provides a detailed drift analysis of the forecasting model across the applications. Finally, Subsection~\ref{subsection: CPU_model} presents the performance of the CPU model that was trained on original TPS values to predict the resource (CPU) across the cloud applications.

\subsection{Performance Analysis of Proposed Two-stage Forecasting Model}
\label{subsection: Forecasting Model Performance}

This section presents a detailed experimental analysis of the newly developed two-stage integrated forecasting system for the given ten cloud applications. The performance of the proposed forecasting model was evaluated using several metrics, including Mean Absolute Error (MAE), Root Mean Square Error (RMSE), Symmetric Mean Absolute Percentage Error (SMAPE), Relative Mean Underestimation Error (rMUE), Relative Mean Overestimation Error (rMOE), the coefficient of determination ($R^2$), and the average forecasting runtime in milliseconds. The XGBoost hyperparameter configuration used in the proposed forecasting model is first presented, followed by the forecasting performance results summarized in Table~\ref{tab:Forecasting_Model_performance}. The first column lists the cloud applications, while the second column indicates the rolling maximum window size (Win.) used for CPU resource smoothing. The remaining columns report the corresponding evaluation metrics.

Prior to evaluating the forecasting performance, the XGBoost regression model was configured using a fixed set of hyperparameters for this experiment. The model employed 300 boosting estimators with a maximum tree depth of 6 and a learning rate of 0.05, providing a balance between model complexity and generalization capability. To mitigate overfitting, both the row subsampling ratio (subsample) and the feature subsampling ratio (colsample\_bytree) were set to 0.8. The regression objective was specified as reg:squarederror. Furthermore, the histogram-based tree construction algorithm (tree\_method = hist) was employed to improve computational efficiency, while parallel execution across all available CPU cores (n\_jobs = -1) reduced the training time. A fixed random seed (random\_state = 42) was used to ensure reproducibility of the experimental results. These parameters come from the Python implementations. The complete hyperparameter configuration is summarized in Table~\ref{tab:xgb_exp1}.

\begin{table}[ht]
\centering
\caption{XGBoost hyperparameter configuration used in the proposed two-stage forecasting model.}
\label{tab:xgb_exp1}
\begin{tabular}{lc}
\hline
Hyperparameter & Value \\
\hline
Random State (random\_state) & 42 \\
Number of Estimators (n\_estimators) & 300 \\
Maximum Tree Depth (max\_depth) & 6 \\
Learning Rate (learning\_rate) & 0.05 \\
Row Subsampling (subsample) & 0.8 \\
Feature Subsampling (colsample\_bytree) & 0.8 \\
Objective Function  & reg:squarederror \\
Tree Construction Method (tree\_method) & hist \\
Parallel Processing (n\_jobs) &  $-1$\\
\hline
\end{tabular}
\end{table}

The results from Table~\ref{tab:Forecasting_Model_performance} suggest that the proposed two-stage forecasting model achieved satisfactory resource (CPU) prediction accuracy for the given cloud applications. The MAE values remained below 1.5 across eight applications, except for application B1, which exhibited a larger prediction error due to its higher workload variability. The lowest MAE was observed for application D5 (0.0005), followed by D4 (0.0011) and D1 (0.0020), which indicates the capability of the proposed model to accurately capture relatively stable workload patterns. Similarly, RMSE values followed a trend consistent with MAE, suggesting that large forecasting deviations were effectively controlled for most applications.

\begin{table*}[ht]
\centering
\caption{Prediction accuracy and computational performance of the proposed  two-stage forecasting model across ten cloud applications}
\label{tab:Forecasting_Model_performance}
\footnotesize
\begin{tabular}{lcccccccc}
\hline
App.& Win.&MAE & RMSE & SMAPE (\%) & rMUE & rMOE & R$^2$ & $RT_{avg}$ (ms) \\
\hline
A& 6 & 0.7372 & 1.1866 & 3.5697  & 0.0222 & 0.0129 & 0.9185 & 186.90 \\
B1& 4 & 6.4567 & 10.9866 & 4.9595 & 0.0332 & 0.0148 & 0.7843 & 181.91 \\
B2& 15 & 0.2784 & 0.4674 & 5.9456 & 0.0145 & 0.0481 & 0.7994 & 184.76 \\
B3& 9 & 0.4631 & 0.7766 & 5.4935 & 0.0303 & 0.0235 & 0.6398 & 232.44 \\
C& 13 & 1.3694 & 2.8348 & 6.3037 & 0.0155 & 0.0579 & 0.7538 & 178.50 \\
D1& 15 & 0.0020 & 0.0033 & 4.0781 & 0.0301 & 0.0090 & 0.6326 & 197.56 \\
D2& 14 & 0.0107 & 0.0236 & 25.4845 & 0.0268 & 0.4832 & 0.6185 & 193.35 \\
D3& 11 & 0.1131 & 0.2430 & 10.9928 & 0.0130 & 0.1258 & 0.6945 & 263.70 \\
D4& 15 & 0.0011 & 0.0023 & 45.7334 & 0.0414 & 1.3155 & 0.6370 & 193.34 \\
D5& 15 & 0.0005 & 0.0008 & 12.2366 & 0.0202 & 0.1256 & 0.6924 & 196.02 \\
\hline
\end{tabular}
\end{table*}

Investigating the effects of rolling maximum-window selection for resource (CPU) smoothing is also crucial for given cloud applications. For all applications, the rolling maximum window size ranged from 4 to 15.  This illustrates how varying workload characteristics necessitate varying degrees of smoothing depending on CPU utilization. With comparatively small window sizes (4 and 6), Applications B1 and A achieved ideal performance, indicating the presence of short-term workload changes. On the other hand, applications B2, D1, D4, and D5 needed larger windows of 15, suggesting that more smoothing was helpful for reducing noise and identifying long-term workload trends. These results confirm that the suggested pre-processing mechanism is adaptable.

In terms of percentage-based forecasting accuracy, we selected the SMAPE metric as given in Equation~\eqref{eq:smape}. The SMAPE values remained below 11\% for most of the applications which confirms the robustness of the forecasting framework under different workload scales. Applications A, B1, B2, B3, C, and D1 achieved SMAPE values below 7\%, indicating highly accurate relative predictions. However, applications D2 and D4 recorded significantly larger SMAPE values of 25.48\% and 45.73\%, respectively. Although D4 exhibits the largest SMAPE (45.73\%), its absolute MAE is only 0.0011. Therefore, the large percentage error is primarily caused by the extremely small CPU utilization values rather than poor forecasting accuracy.

From Equations~\eqref{eq:mue} and \eqref{eq:moe}, we computed the relative mean underestimation error (rMUE) and relative mean overestimation error (rMOE) for all the given applications using the proposed two-stage integrated forecasting model. These metrics help us to understand the behavior of the proposed forecasting model in terms of overestimation and underestimation. The rMUE remained consistently low across all applications, ranging between 0.0130 and 0.0414. This indicates that the model rarely produced substantial under-predictions. Similarly, the rMOE values were relatively small for most applications, demonstrating balanced forecasting behavior.

Nevertheless, applications D2 and D4 displayed comparatively higher overestimation tendencies with rMOE values of 0.4832 and 1.3155. Such behavior suggests that the forecasting model overestimates workloads when CPU utilization levels are extremely low. From a cloud resource management perspective, moderate overestimation is generally preferable to underestimation because it reduces the risk of service-level agreement (SLA) violations caused by insufficient resource allocation.

Furthermore, the values of the coefficient of determination ($R^2$) varied between 0.6185 and 0.9185 for all cloud applications. Application A had the greatest value ($R^2 = 0.9185$), meaning that the suggested forecasting model accounted for over 91\% of the workload variance. The majority of cloud applications showed a good correlation between expected and actual CPU consumption patterns with $R^2$ values above 0.69. Lower $R^2$ values for D2, D4, and B3 indicate the existence of erratic, hard-to-predict workload changes.

In addition to prediction accuracy, the computational efficiency of the proposed forecasting model was evaluated using Equation~\eqref{eq:runtime}, which computes the average runtime required to generate resource (CPU) predictions. The average runtime ranged from 178.50 ms to 263.70 ms, with a mean runtime of approximately 201 ms across all applications. These results indicate that the newly developed forecasting model can generate predictions in less than one quarter of a second, making it suitable for real-time cloud resource provisioning and auto-scaling applications. Although application D3 required the highest runtime (263.70 ms), the computational overhead remains sufficiently low for practical deployment in dynamic cloud environments.

The presented forecasting model facilitates the proactive resource provisioning in private cloud systems due to its low sub-second runtime. Precise CPU estimates reduce SLA violations and increase resource utilization efficiency by enabling auto-scaling techniques to distribute resources prior to workload peaks. Since application $A$ offers a sample example of the forecasting behavior seen throughout the examined cloud applications, only its forecasting findings are graphically given for the sake of conciseness.

\begin{figure*}[ht]
	\centering
	\includegraphics[width=0.9\linewidth]{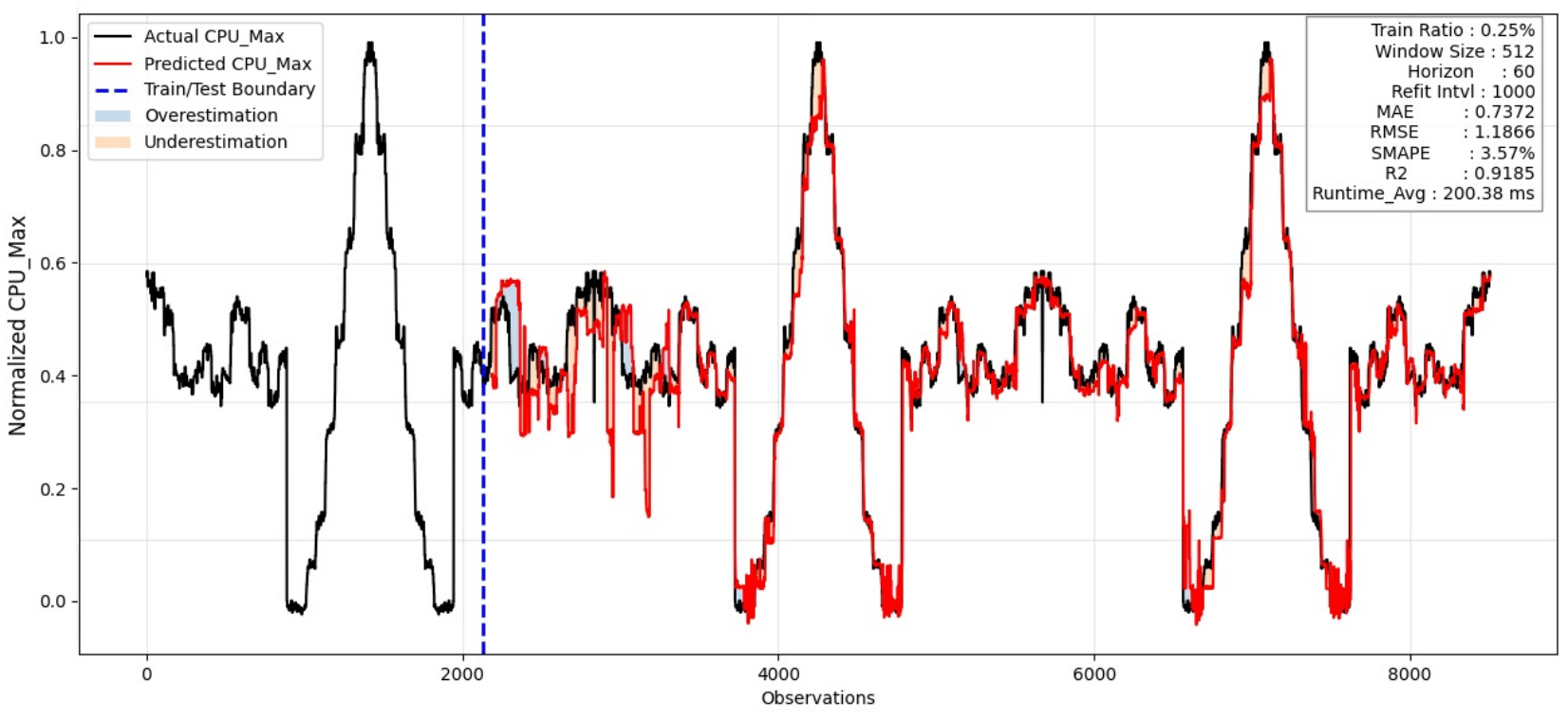}
	\caption{Performance analysis of the proposed two-stage integrated forecasting model for the application $A$, where a forecasting horizon of 60 corresponds to a 1-hour-ahead resource prediction.}
	\label{fig:Forecasting_App_A}
\end{figure*}

Figure~\ref{fig:Forecasting_App_A} presents the forecasting performance of the proposed two-stage integrated forecasting model for application $A$. The black curve represents the actual normalized $CPU\_Max$ values, whereas the red curve corresponds to the forecasts generated by the proposed model. The vertical blue dashed line indicates the transition from the training phase to the testing phase, where the initial 25\% of the observations were used for model training, and the remaining 75\% were reserved for walk-forward evaluation. The forecasting framework employs a rolling window of 512 observations to predict future TPS values over a 60-step forecasting horizon, which the CPU prediction model subsequently uses to estimate future $CPU\_Max$ utilization.

A visual examination of the testing area shows that the expected and actual CPU workload match quite well. Abrupt workload shifts, high-utilization peaks, and low-utilization periods are only a few of the long-term workload dynamics that the suggested framework effectively captures. Specifically, during periods of relatively low CPU activity, the model maintains constant performance while precisely tracking the significant utilization spikes around observations 4,200 and 7,100. The shaded overestimation and underestimation regions' modest extent further suggests that predicting discrepancies is minimal for the majority of the evaluation period.

These findings are further supported by the quantitative results shown in the Figure~\ref{fig:Forecasting_App_A} for application A. With an MAE of 0.7372, an RMSE of 1.1866, an SMAPE of 3.57\%, and an $R^2$ value of 0.9185, the proposed model effectively explained over 91\% of the variability in CPU use. The high coefficient of determination and low forecasting errors indicate that the suggested framework effectively preserves workload characteristics over the forecasting horizon. These findings show that the proposed forecasting model preserves the temporal features of CPU usage dynamics while maintaining forecasting accuracy over long prediction horizons.

Overall, the experimental findings show that the newly developed two-stage forecasting model strikes a good balance between operational robustness, computational efficiency, and forecasting accuracy. Across a variety of cloud applications, the model continuously maintained low forecasting errors, high explanatory power, and quick execution times. These capabilities make the proposed forecasting model efficient for proactive resource provisioning, intelligent auto-scaling, and real-time CPU workload predictions in dynamic cloud computing settings.

\subsection{Drift Analysis of Forecasting Model}
\label{subsection: Detailed Drift Analysis}
Although the overall forecasting accuracy provides a global assessment of model performance, it does not reveal how prediction errors evolve as the forecasting horizon increases. In recursive multi-step forecasting, prediction errors generated at earlier steps are propagated to subsequent predictions, which may lead to the accumulation of forecasting drift over the time. Therefore, a drift analysis was conducted to evaluate the stability and robustness of the proposed two-stage forecasting system across the forecasting horizon.



The following anonymized TPS features were examined for this purpose: \emph{Customer Service Request 1} (CSR1), \emph{Customer Service Request 2} (CSR2), \emph{Customer Service Request 3} (CSR3), \emph{Customer Service Request 4} (CSR4), and \emph{Customer Service Request 5}(CSR5). The original service identifiers were anonymized to preserve the confidentiality of customer applications and cloud services while retaining the statistical characteristics required for forecasting and resource estimation analyses.

\begin{figure}[ht]
	\centering
	\includegraphics[width=\linewidth]{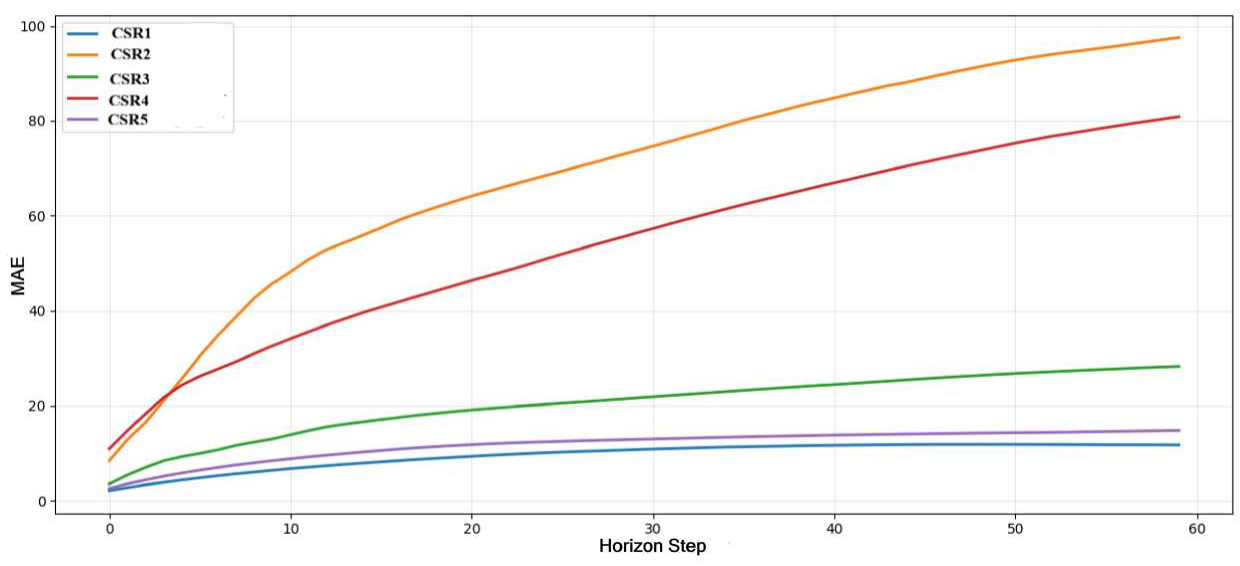}
	\caption{Evolution of forecasting drift over a 60-step (0-59) prediction horizon measured by horizon-wise MAE. Error increases progressively for all TPS traffic streams, with CSR2 and CSR4 exhibiting the largest drift, while CSR1, CSR3, and CSR5 maintain comparatively stable long-term prediction performance.}
	\label{fig:Drift_Line}
\end{figure}

Figure~\ref{fig:Drift_Line} illustrates the evolution of forecasting drift across a 60-step prediction horizon using horizon-wise MAE values, starting from 0 to 59 steps where 0 represents the first predicted value. The figure provides a direct visualization of recursive error propagation and highlights the long-term stability characteristics of each TPS traffic stream. For every TPS feature, there is a monotonic rise in predicting error as shown in Figure~\ref{fig:Drift_Line}, supporting the anticipated build-up of uncertainty in recursive multi-step forecasting. CRS2 and CSR4 show the sharpest development, with MAE values rising steadily along the forecasting horizon and reaching the maximum error magnitudes at longer prediction intervals. This trend suggests that these traffic streams have more intricate temporal dynamics and are more susceptible to recurrent forecasting errors. On the other hand, CSR1, CSR3, CSR5 show significantly slower drift increase. Their MAE curves show better long-range predictability and higher temporal consistency as they progressively rise and stabilize over longer time periods.

Furthermore, none of the TPS features show an abrupt divergence or an exponential rise in error. Rather, the drift curves remain smooth and roughly linear, suggesting that prediction uncertainty accumulates in a regulated way. This behavior indicates that the forecasting model does not exhibit catastrophic recursive error propagation and remains stable throughout the entire prediction horizon. Therefore, the drift growth study confirms the usefulness of the proposed TPS forecasting component as an input generator for downstream CPU resource prediction and demonstrates its resilience.



\subsection{Standalone Evaluation of the CPU Prediction Model}
\label{subsection: CPU_model}
Unlike Section~\ref{subsection: Forecasting Model Performance}, which evaluates the performance of the complete two-stage forecasting system, this section evaluates the CPU prediction model independently. The objective is to assess the intrinsic prediction accuracy and computational efficiency of the CPU model without the influence of errors propagated from the TPS forecasting stage, thereby distinguishing the contribution of the CPU prediction component from that of the overall forecasting system.

\begin{table}[ht]
\centering
\caption{XGBoost hyperparameter configuration}
\label{tab:xgb_cpu_params}
\begin{tabular}{lc}
\hline
Hyperparameter & Value \\
\hline
Random State (random\_state)& 42 \\
Number of Estimators (n\_estimators) & 300 \\
Maximum Tree Depth (max\_depth) & 4 \\
Learning Rate (learning\_rate) & 0.05 \\
Row Subsampling (subsample) & 0.8 \\
Feature Subsampling (colsample\_bytree) & 0.8 \\
Objective Function  & reg:squarederror \\
\hline
\end{tabular}
\end{table}

For the standalone CPU prediction experiment, the XGBoost regression model was configured using a separate set of hyperparameters to evaluate the intrinsic prediction capability of the CPU model. The model employed 300 boosting estimators with a maximum tree depth of 4 and a learning rate of $0.05$. To improve generalization and reduce overfitting, both the row subsampling ratio ($subsample$) and the feature subsampling ratio ($colsample\_bytree$) were fixed at 0.8. The regression objective was specified as reg:squarederror, and a fixed random seed of 42 was used to ensure reproducibility of the experimental results. The complete hyperparameter configuration is presented in Table~\ref{tab:xgb_cpu_params}. Unlike to Experiment~\ref{subsection: Forecasting Model Performance}, where the XGBoost model employed parallel processing ($n\_jobs$) as shown in Table~\ref{tab:xgb_exp1}, Experiment~\ref{subsection: CPU_model} (Table~\ref{tab:xgb_cpu_params}) focuses solely on CPU resource prediction without the TPS forecasting stage. Consequently, parallel processing ($n\_jobs$) was not utilized, as the experiment was designed to evaluate only the CPU prediction model.

The performance of the resource prediction (CPU) model was evaluated using multiple error metrics, including MAE, RMSE, SMAPE, rMUE, rMOE, and the coefficient of determination ($R^2$), across nine cloud applications. In addition, the average response time ($RT_{avg}$) was measured to assess the computational efficiency of the prediction process. The results are summarized in Table~\ref{tab:CPU_Model_performance}.

\begin{table*}[ht]
\centering
\caption{Performance of CPU model for resource prediction using original TPS features for given cloud applications}
\label{tab:CPU_Model_performance}
\footnotesize
\begin{tabular}{lcccccccc}
\hline
App. & Win&MAE & RMSE & SMAPE (\%) & rMUE  & rMOE & R$^2$ & $RT_{avg}$(ms) \\
\hline
A  & 6&0.216 & 0.409 & 1.0400 & 0.0099 & 0.0109 & 0.9903 & 1.138591 \\
B1 & 4&2.232 & 3.528 & 1.7866 & 0.0158 & 0.0183 & 0.9776 & 4.530044 \\
B2 & 15&0.111 & 0.216 & 2.3929 & 0.0224 & 0.0249 & 0.9567 & 7.673111 \\
B3 &9 &0.242 & 0.377 & 2.9342 & 0.0265 & 0.0305 & 0.9145 & 5.336268 \\

C & 13&0.553 & 1.123 & 2.4841 & 0.0258 & 0.0269 & 0.9611 & 1.042598\\

D1 &15 &0.001 & 0.001 & 1.8528 & 0.0183 & 0.0189 & 0.9438 & 1.631443 \\
D2 & 14 &0.002 & 0.005 & 6.2378 & 0.0345 & 0.0847 & 0.9821 & 0.842303 \\
D3 & 11 &0.026 & 0.061 & 3.1652 & 0.0259 & 0.0346 & 0.9809 & 0.836380 \\
D4 &15 &0.000 & 0.001 & 19.7295 & 0.1064 & 0.3074 & 0.9737 & 0.787613 \\
D5 & 15& 0.000 & 0.000 & 7.4061 & 0.0546 & 0.0807 & 0.9328 & 0.771310 \\
\hline
\end{tabular}
\end{table*}

The consistently high $R^2$ values, ranging from 0.9145 to 0.9903 across the evaluated cloud applications, indicate that the CPU model achieved high prediction accuracy. With an $R^2$ value of 0.9903, a low MAE of 0.216, and a SMAPE of only 1.04\%, application A had the best predictive performance, demonstrating outstanding agreement between the projected and real CPU usage statistics. Applications D2 and D3 also demonstrated outstanding predictive skills, sustaining relatively low prediction errors while obtaining $R^2$ values above 0.98.

The prediction performance of B1, despite having the highest absolute error of any application (MAE = 2.232 and RMSE = 3.528), B1 nevertheless had a good coefficient of determination ($R^2 = 0.9776$), suggesting that the model was successful in capturing the general patterns in CPU consumption. While the SMAPE values of 2.39\% and 2.93\%, respectively, show acceptable relative prediction accuracy, the MAE values for B2 and B3 stayed below 0.25.

An intriguing finding on the pod D applications. The MAE and RMSE for Applications D4 and D5 were very low, indicating negligible absolute prediction errors. Nonetheless, the SMAPE, rMUE, and rMOE values for these applications were significantly greater. Specifically, D4 exhibited the highest SMAPE (19.73\%), rMUE (10.64\%), and rMOE (30.74\%) among all evaluated applications. This behavior can be attributed to the relatively small magnitude of the actual CPU utilization values in these workloads, where even minor prediction deviations can result in large percentage-based errors. Therefore, the elevated relative error metrics for D4 and D5 should be interpreted in the context of their very low absolute resource utilization levels rather than as indicators of poor model performance.

Additionally, Applications A, B1, B2, B3, D1, and D3 have rMUE values below 3\%, and the associated rMOE is also below 4\%. These findings show that the model regularly generates predictions with modest overestimation and underestimation behavior that are near the observed CPU usage values. The model's appropriateness for near real-time resource prediction in cloud settings is demonstrated by the average predicted reaction time, which spans from 0.77 ms to 7.67 ms across all applications. B2 needed the longest average reaction time of 7.67 ms, whereas applications D4 and D5 had the lowest response times of 0.79 ms and 0.77 ms, respectively. However, every prediction time stays within a few milliseconds, indicating that the CPU prediction model is lightweight.

In conclusion, the findings show that the CPU prediction model generates highly accurate and computationally efficient resource projections across a wide range of cloud applications. The model's continuously high $R^2$ values, low absolute errors, and millisecond-level forecast latency show that it can enable proactive resource management and dynamic scaling decisions in cloud computing.

\section{Discussion}
\label{sec:discussion}
The newly developed two-stage integrated forecasting model successfully supports proactive cloud resource prediction, as the experimental evaluation confirms. The proposed forecasting methodology continuously attained excellent prediction accuracy while preserving computational efficiency across a wide range of cloud applications by fusing customer service request (TPS) forecasting with CPU workload prediction. The conventional resource prediction approaches directly estimate future CPU workload using timestamps or historical data; the proposed forecasting model first predicts future customer service requests (TPS) behavior and subsequently estimates resource (CPU) demand. This hierarchical modeling strategy enables the forecasting model to capture the causal relationship between workload dynamics and CPU utilization more effectively.

One of the major findings of this study is the consistently low forecasting error across most applications. The obtained MAE, RMSE, and SMAPE values indicate that the proposed forecasting model accurately models both high-load and moderate-load applications despite the workload characteristics. The high coefficient of determination ($R^2$) observed for nearly all applications further confirms that the proposed forecasting model successfully captures the TPS variability of cloud workloads. Applications exhibiting lower prediction accuracy such as B3, D2, and D4 that are correspond to workloads with highly irregular CPU utilization patterns or extremely low CPU values, making them inherently more difficult to predict. Nevertheless, even in these challenging cases, the forecasting performance remained stable without catastrophic degradation.

Additional proof of the proposed forecasting strategy's resilience is provided by the drift study. Prediction errors in recursive multi-step forecasting inevitably accumulate as the forecasting horizon lengthens. Rather than exponential drift accumulation, the slow rise in MAE indicates regulated error propagation as shown in Figure~\ref{fig:Drift_Line}. In practical cloud resource provisioning, where auto-scaling decisions are directly impacted by prediction reliability over several future intervals, this feature is especially crucial.

A comparison of the results presented in Tables~\ref{tab:Forecasting_Model_performance} and \ref{tab:CPU_Model_performance} provides valuable insight into the impact of workload forecasting on the overall prediction framework. It should be noted that these tables correspond to two different experimental settings. In the first experiment, the proposed two-stage integrated system uses forecasted TPS values generated by the forecasting model as inputs to the CPU prediction model. Consequently, the final CPU prediction accuracy is influenced by both the forecasting uncertainty and the regression error, leading to relatively lower prediction performance. In contrast, the second experiment evaluates the CPU prediction model using the original TPS values, thereby isolating the performance of the resource prediction model from forecasting errors. As expected, this configuration achieves substantially lower MAE and RMSE values and higher $R^2$ values across all cloud applications. The observed performance gap between the two experiments quantifies the effect of forecasting uncertainty on downstream resource prediction. Despite relying on predicted TPS values, the integrated forecasting system still maintains satisfactory prediction accuracy and sub-second execution time, demonstrating that the proposed forecasting component introduces only a limited degradation in CPU prediction performance. These findings confirm the robustness of the two-stage architecture and indicate that the forecasting errors are sufficiently controlled to support practical proactive resource provisioning and auto-scaling in dynamic cloud environments.

The proposed forecasting model's computational efficiency is another noteworthy result. While the CPU prediction model itself runs in a matter of milliseconds, the forecasting stage takes just about 200 ms to construct a full 60-step prediction horizon. The two-stage forecasting system enables dynamic resource allocation and real-time cloud monitoring systems without causing appreciable scheduling delays due to its low computational overhead. As a result, the proposed forecasting approach can be implemented in real-world cloud settings where prediction latency is crucial.

The proposed forecasting approach has a number of benefits over several current cloud resource prediction techniques, which mostly concentrate on one-step forecasting or direct CPU estimation. First, the two-stage architecture improves interpretability and forecasting flexibility by explicitly modeling workload evolution before resource prediction. Second, the incorporation of drift analysis offers a thorough assessment of long-horizon forecasting stability, which is frequently disregarded in earlier research. Third, the proposed approach takes into account forecasting robustness, computing efficiency, and prediction accuracy all at once, which increases its applicability to real-world cloud resource management situations.

Despite these encouraging results, several limitations remain. The experimental evaluation is based on TPS-driven workload data collected from a private cloud environment and focuses exclusively on CPU resource prediction. Other critical cloud resources, such as memory, storage, network bandwidth, and energy consumption, were not considered in this study. Furthermore, the forecasting system was evaluated using historical workload traces under relatively stable operational conditions, and the dataset size and workload diversity were limited.

In future work, we plan to extend the analysis using larger-scale datasets collected over longer time periods with more diverse and intensive workload patterns. Adaptive online learning mechanisms may be required to maintain prediction accuracy under unexpected workload anomalies, hardware failures, or concept drift during long-term deployments. However, a key limitation of the current expanding-window retraining strategy is its computational overhead. As the number of observations increases over time, the retraining cost of the forecasting model also increases due to the continuously growing training dataset. Future research will therefore focus on extending the proposed approach to multivariate resource prediction, simultaneously modeling CPU, memory, storage, and network utilization. The robustness of long-term forecasting can be further improved by incorporating transformer-based sequence models, continual learning strategies, and advanced drift adaptation techniques. Additionally, integrating the proposed forecasting framework with reinforcement learning-based auto-scaling mechanisms represents a promising direction toward fully autonomous cloud resource management systems.

\section{Conclusion}
\label{sec:conclusion}
This study presented a two-stage integrated forecasting model for proactive resource (CPU) prediction in cloud computing environments. The conventional approaches directly forecast future CPU workload from historical resource traces. At the same time, the proposed forecasting model explicitly models the relationship between customer service demand and resource consumption by first forecasting future Transactions Per Second (TPS) and subsequently estimating downstream CPU workload.
The proposed technique uses the XGBoost model in both components (forecasting and resource prediction). It includes adaptive online retraining via an expanding-window strategy to continually adapt to changing workload patterns while keeping computational overhead low.

The proposed approach was extensively evaluated using real-world workload traces collected from a private cloud infrastructure comprising ten cloud applications. The two-stage integrated forecasting model achieved accurate CPU forecasting across various workloads in cloud applications, with a median Symmetric Mean Absolute Percentage Error (SMAPE) value of $5.9\%$ for the complete forecasting model and a $R^2$ value between $0.9145 - 0.9903$ for the CPU estimation module. The average forecasting latency for the integrated forecasting pipeline was less than 264 ms and less than 1 ms for CPU estimate, showing that it is suitable for real-time deployment in cloud auto-scaling systems. Furthermore, horizon-wise drift study revealed that recursive forecasting errors were comparatively stable across a 60-step prediction horizon for $3$ out of $5$ features, demonstrating the feasibility of the proposed cascaded forecasting architecture.

For cloud resource management, the proposed forecasting approach provides a number of useful benefits. Cloud providers can anticipate workload changes earlier than with traditional direct forecasting methods by predicting customer demand before estimating resource requirements. This enables proactive resource provisioning, reduces Service Level Agreement (SLA) violations, improves resource utilization, and lowers operating costs. Additionally, XGBoost offers a computationally efficient alternative to deep learning-based forecasting models.


\section*{Authorship contribution}
\addcontentsline{toc}{section}{Authors' Contributions}
\textbf{Ashir Javeed:} Conceptualization, Methodology, Validation, Writing – original draft.
\textbf{Anton Borg:} Methodology, Writing – review \& editing.
\textbf{Håkan Grahn:} Methodology, Validation, Supervision, Resources, Writing – review \& editing.
\textbf{Lars Lundberg:} Methodology, Supervision, Writing – review \& editing.
\textbf{Dhyey Patel:} Resources, Data Curation, Writing – review \& editing.
\textbf{Sogand Shirinbab:} Resources, Data Curation, Writing – review \& editing.

\section*{Funding}
This work is funded in part by the project ``Green Clouds - Load Prediction and Optimization in Private Cloud Systems'', funded by the Knowledge Foundation (grant: 20220215) in Sweden.

\section*{Acknowledgements}
The authors thank Ericsson AB, Karlskrona, Sweden for providing data and technical support for this work.

\section*{Data availability}
\addcontentsline{toc}{section}{Availability of Data and Materials}
The data used in the study is confidential.

\section*{Conflict of interest/Competing interests}
\addcontentsline{toc}{section}{Declaration of competing interest}
The authors declare that they have no known competing financial interests or personal relationships that could have appeared to influence the work reported in this paper.


\bibliographystyle{elsarticle-num}
\bibliography{bibliography}

\end{document}